\pdfoutput=1

\documentclass[11pt]{article}

\usepackage[]{naacl2021}

\usepackage{times}
\usepackage{latexsym}
\usepackage{amssymb}
\usepackage{graphics}
\usepackage{enumitem}
\usepackage{textcomp}
\usepackage{amsmath}
\usepackage{devanagari}
\usepackage{booktabs}

\usepackage[T1]{fontenc}

\usepackage[utf8]{inputenc}

\usepackage{microtype}

\title{Representing Numbers in NLP: a Survey and a Vision}

\author{
Avijit Thawani \and Jay Pujara \and Pedro Szekely \and Filip Ilievski \\
University of Southern California \\ Information Sciences Institute
\\ \texttt{\{thawani,jpujara,pszekely,ilievski\}@isi.edu}
}

\begin{document}
\maketitle
\begin{abstract}
NLP systems rarely give special consideration to numbers found in text. This starkly contrasts with the consensus in neuroscience that, in the brain, numbers are represented differently from words. We arrange recent NLP work on numeracy into a comprehensive taxonomy of tasks and methods. We break down the subjective notion of numeracy into 7 subtasks, arranged along two dimensions: granularity (exact vs approximate) and units (abstract vs grounded). We analyze the myriad representational choices made by 18 previously published number encoders and decoders. We synthesize best practices for representing numbers in text and articulate a vision for holistic numeracy in NLP, comprised of design trade-offs and a unified evaluation.

\end{abstract}
\section{Introduction}
\label{intro}

Numbers are an integral part of text. To understand a simple sentence like \textit{I woke up at 11}, we need not just literacy but also numeracy. We must decode the string \textit{11} to the quantity {\ttfamily 11} and infer {\ttfamily 11} to denote a time of the day, probably {\ttfamily 11 a.m.} We need commonsense to reason that 11 a.m. is quite late in the morning.
This interpretation of {\ttfamily 11} is strongly contextual, as \textit{I earn \$11 per month} evokes different units and value expectations.
Note how the semantics remains the same for both sentences if {\ttfamily 11} was replaced by {\ttfamily 10}, i.e., the context is tolerant to some variability.


\textbf{Numbers are everywhere}. 
Reasoning with quantities and counts is crucial to understanding the world. Evolutionary learning has given numerical cognition skills 
to several animals, including human beings \cite{dehaene2011number}.
Our ancient ancestors furthered numeracy by developing multiple number systems, similar to but independent from the evolution of languages. 
Numeracy is an essential skill for language understanding, since numbers are often interspersed in text:
the $6$ million pages in English Wikipedia have over $150$ million numbers.


\textbf{Numbers are neglected.} In NLP, however, numbers are either filtered out explicitly during preprocessing \cite{graff2003english}, or treated the same as words, often collapsing them into an {\ttfamily UNK} token. Subword tokenization approaches like BPE \cite{sennrich-etal-2016-neural} and WordPiece \cite{wu2016googles} instead retain numbers, but split them into arbitrary tokens, for example {\ttfamily 1234} might be split into two tokens as {\ttfamily 12-34} or {\ttfamily 123-4} or {\ttfamily 1-234}. 


Recent work has shown that these are suboptimal number representations \cite{wallace-etal-2019-nlp,zhang2020scale}. On the DROP Question Answering benchmark, BERT performs five times worse when the answer is a number instead of a span of text \cite{dua-etal-2019-drop}. Relatively simple strategies like switching from subword to char-level tokenization \cite{geva-etal-2020-injecting}, or from decimal to scientific notation \cite{zhang2020scale} already boost performance. Such results warrant a deeper study into the best number representations.

\textbf{Numbers are important}. Given the ubiquity of numbers and their fundamental differences with words, enabling NLP systems to represent them effectively is beneficial for domains like scientific articles \cite{riedel2018} and financial documents \cite{chen-etal-2019-numeracy,jiang-etal-2020-learning}. Number understanding is also useful to detect sarcasm \cite{dubey-etal-2019-numbers} and to model dialogues involving price negotiations \cite{chawla2020bert}.

Recent NLP progress towards numeracy has been sporadic but encouraging. In this paper, we survey prior work and highlight the kind of numeracy targeted (e.g., arithmetic, measurement, numeration) as well as the kind of representation used (e.g., value embeddings, DigitRNNs). We provide the first NLP-centric taxonomy of numeracy tasks (Section \ref{tasks}) and of number representations (Section \ref{methods}) for the reader to succinctly comprehend the challenge posed by numeracy. We synthesize key takeaways (Section \ref{takeaways}) and propose a unifying vision for future research (Section \ref{roadmap}).

\begin{table*}
    \centering
    \begin{tabular}{l|cc|c}
    \toprule
    & \multicolumn{2}{c|}{\textbf{Benchmarking or Probing Tasks}} & \textbf{ Downstream } \\ 
    & \textbf{ Abstract } & \textbf{ Grounded } & \textbf{ Applications } \\ 
    \midrule
    \textbf{Exact} & Simple Arithmetic (2+3=5) & AWP (2 balls + 3 balls = 5 balls), & Question Answering, 
    \\
    & & Exact Facts (birds have two legs) & Science Problems
    \\
    \midrule
    \textbf{Approx} & Numeration (`2' = 2.0), & Measurement (dogs weigh 50 lbs), & Sarcasm Detection, 
    \\
    & Magnitude (`2' < `5') & Numerical Language Modeling & Numeral Categorization \\
    \bottomrule
    \end{tabular}
    \caption{Seven numeracy tasks, arranged along the axes of (rows) granularity - exact vs approximate, and (columns) units - abstract vs grounded. We also list downstream applications requiring a similar granularity of numeracy. 
    }
    \label{tab:tasks}
\end{table*}

\section{Tasks}

\label{tasks}

There are several different aspects of numeracy. The DROP dataset alone offers a wide variety of numeric reasoning questions such as retrieval-based (\textit{How many yards did Brady run?}), count-based (\textit{How many goals were scored?} given a comprehension describing multiple goals), and simple arithmetic (\textit{How many years after event 1 did event 2 occur?} given dates of both events). 
Besides downstream applications, there have also been probing experiments to evaluate whether NLP models can decode numbers from strings (e.g., \textit{19} to {\ttfamily 19.0}), or estimate quantities (e.g., \textit{how tall are lions?}). 

Such a diverse range of abilities are usually all referred to collectively as \textit{numeracy}, which gives rise to confusion. We limit this abuse of terminology and provide a neat taxonomy for arranging the different tasks proposed under numeracy.


\subsection{Our Taxonomy of Tasks}
\label{tasks:taxonomy}

Drawing from work in cognitive science \cite{feigenson2004core}, we propose the following two dimensions to organize tasks within numeracy:

\begin{enumerate}[wide, labelwidth=0pt, labelindent=5pt]
    \item \textbf{Granularity}: whether the encoding of the number is (1) exact, e.g., \textit{birds have {\ttfamily two} legs}, or (2) approximate, e.g., \textit{Jon is about {\ttfamily 180} cms tall}. 
    \item \textbf{Units}: whether the numbers are (1) abstract, e.g., 2+3=5, or (2) grounded, e.g., 2 apples + 3 apples = 5 apples. While abstract mathematical tasks are easy to probe and create artificial datasets for, numbers grounded in units are challenging since they need to be understood in the context of words.
\end{enumerate}

\subsection{Survey of Existing Tasks}
\label{tasks:existing}

We now describe 7 standard numeracy tasks, arranged according to our taxonomy in Table \ref{tab:tasks}. We defer discussion on downstream tasks (right-most column in the table) as well as miscellaneous numeracy-related tasks (such as counting) to \ref{tasks:misc}. 

\textbf{Simple Arithmetic} is the task of addition, subtraction, etc. over numbers alone. It is convenient to create synthetic datasets involving such math operations for both masked \cite{geva-etal-2020-injecting} and causal language models (GPT-3 \citealt{brown2020language}).

\textbf{Numeration} or Decoding refers to the task of mapping a string form to its numeric value, e.g., \textit{19} to {\ttfamily 19.0}. Within NLP, this task is set up as a linear regressor probe over a (frozen) representation of the string. Numeration has been probed for in static word embeddings \cite{naik-etal-2019-exploring}, contextualized language models \cite{wallace-etal-2019-nlp}, and multilingual number words, e.g., \textit{nineteen} or \textit{dix-neuf} \cite{johnson-etal-2020-probing}.

\textbf{Magnitude Comparison} is the ability to tell which of two (or more) numbers is larger. For language models, this has been probed in an argmax setup (choose the largest of five numbers) as well as a binary classification task, e.g., given \textit{23} and \textit{32}, pick the label {\ttfamily 1} to indicate that \textit{32} > \textit{23} \cite{naik-etal-2019-exploring,wallace-etal-2019-nlp}.

\textbf{Arithmetic Word Problems (AWP)} are the grounded version of simple arithmetic that we find in school textbooks, e.g., \emph{Mary had two cookies. She gave one away. How many does she have left?} There exist several NLP datasets on math word problems \cite{amini-etal-2019-mathqa,saxton2018analysing,royAWP,hendrycksmath2021}.

\textbf{Exact Facts} in the context of numeracy involves commonsense knowledge such as \emph{dice have {\ttfamily 6} faces} or \emph{birds have {\ttfamily two} legs}. An approximate sense of quantity would be of little help here since assertions like \emph{dice have {\ttfamily 5} faces} or \emph{birds have {\ttfamily three} legs} are factually incorrect. Two recent datasets for numeric commonsense facts are Numbergame \cite{mishra2020question} and NumerSense \cite{lin2020nummersense}.

\textbf{Measurement Estimation}
is a task in psychology in which subjects are asked to approximately guess measures of objects along certain dimensions, e.g., \emph{number of seeds in a watermelon} or \emph{weight of a telephone} \cite{bullard2004biber}.
VerbPhysics \cite{forbes2017verb} is a benchmark of binary comparisons between physical attributes 
of various objects, e.g., ball $<_{size}$ tiger. DoQ \cite{elazar-etal-2019-large} is a web-extracted dataset of Distributions over Quantities, which can be used as a benchmark for language models' measurement estimation abilities \cite{zhang2020scale}. Lastly, MC-TACO \cite{zhouTACO} is a collection of temporal-specific measurement estimates, e.g., \emph{going for a vacation spans a few days/weeks}.


\textbf{Numerical Language Modeling} 
in its literal sense is not a task but a setup, analogous to masked/causal language modeling for words. Other tasks could be modeled as numeric language modeling, e.g., arithmetic (\emph{2+3=[MASK]}) and measurement estimation (\emph{lions weigh [MASK] pounds}). 
In practice, numerical language modeling refers to the task of making numeric predictions for completing unlabelled, naturally occurring text.

Word predictions in language modeling are typically evaluated with classification metrics such as Accuracy at $K$ or perplexity.
Numeric predictions, on the other hand, are evaluated with regression metrics such as mean absolute error, root mean squared error, or their log-scaled and percentage variants. 

\textbf{Downstream Applications} for numeracy are abound. \citet{dubey-etal-2019-numbers} detect sarcasm in tweets based on numbers.
\citet{chen2020numclaim} identify claims in financial documents using alternative number representations and the auxiliary task of numeral understanding or categorization \cite{chen2018numeral}. Similarly, simple arithmetic and math word problems serve as auxiliary tasks for GenBERT \cite{geva-etal-2020-injecting} towards improving its score on the DROP QA benchmark.

\section{Methods}
\label{methods}
Analogous to our taxonomy of subtasks in the previous section, here we attempt to arrange the wide variety of alternative number representations proposed in recent literature. 
We limit our analysis to methods of encoding (numbers → embeddings) and/or decoding (embeddings → numbers) numbers.
We do not discuss, for example, methods that use symbolic reasoning \cite{andor-etal-2019-giving} or modify activation functions to enhance numeracy \cite{trask2018neural}. 

A typical example of the base architecture could be BERT \cite{devlinbert}, the workhorse of modern NLP. We assume that there exists an independent parallel process of mapping words into embeddings, such as subword tokenization followed by lookup embeddings in BERT. 

\subsection{Our Taxonomy}
\label{methods:taxonomy}
We look at two kinds of representations: string-based and real-based. Real-based representations perform some computation involving the numerical value of the number. The string-based representations instead see numbers in their surface forms; they must assign arbitrary token IDs and look up their embeddings to feed into the architecture.

\subsubsection{String Based}
\label{methods:string}

By default, language models treat numbers as strings, the same as words. However, within string representations, one could tweak simple changes:

\textbf{Notation}: The number 80 could be written in Hindu-Arabic numerals (80), Roman numerals (LXXX), scientific notation (8e1), English words (eighty), or with base 20 as in French (quatre-vingts). 
\citet{nogueira2021investigating} exclusively study the effect of many such notation choices in language models, on the task of simple arithmetic. 

\textbf{Tokenization}: Word level tokenizations are ineffective for numbers, since they are likely to map most numbers to an UNK token, except for a few commonly occuring ones (e.g., 1, 2, 5, 10, 100). Other possibilities are subword tokenizations like BPE and WordPiece, as well as character (or digit) level tokenizations. 
    
\textbf{Pooling}: The pooling dimension of variation springs up after analyzing the effect of tokenization. With subword and character level tokenizations, a single number may now correspond to multiple tokens, e.g., 100 segmented into {\ttfamily 10-0} or {\ttfamily 1-0-0}. Prior work \cite{riedel2018} has argued for using RNNs or CNNs to instead pool the embeddings of these tokens into a single embedding before feeding to the language model. The default way that language models see numbers are the same as words, hence no pooling is applied.

\subsubsection{Real Based}
\label{methods:real}
Real-based number encoders can be expressed as $f:\mathbb{R}\to\mathbb{R}^d$ whereas decoders can be expressed as $g:\mathbb{R}^d\to\mathbb{R}$. Real-based methods proposed in literature can vary on account of direction (whether they encode, decode or both), scale (linear vs log), and discretization (binning vs continuous valued).


\textbf{Direction}: Some proposed methods are encoder-only, e.g., DICE \cite{sundararaman-etal-2020-methods}, while some can be decoder-only, e.g., those requiring sampling from a parameterized distribution \cite{berg-kirkpatrick-spokoyny-2020-empirical}.

\textbf{Scale}: Inspired by cognitive science literature \cite{dehaene2011number}, several methods have attempted to model numbers in the log (instead of linear) scale, i.e., to perform mathematical operations on the logarithm of the number to be represented.
The first operation in a log-scaled $f$ is $log(\cdot)$ and the last operation in a log-scaled $g$ is $exp(\cdot)$.
We discuss more scales in the following subsection, such as the stabilized log scale \cite{jiang-etal-2020-learning} and the learned scale/flow \cite{berg-kirkpatrick-spokoyny-2020-empirical}.


\textbf{Discretization}: Training continuous value functions for a large range of numbers turns out to be practically infeasible \cite{wallace-etal-2019-nlp}. Some real-based methods first bin numbers before learning embeddings for each bin. These bins could be on the linear scale (0-10, 10-20, 20-30, \dots) or the log scale (0.01-0.1, 0.1-1, 1-10, \dots), and the lookup embeddings can be learnt by the regular cross entropy \cite{chen2020numclaim} or dense cross entropy \cite{zhang2020scale}.

\subsection{Survey of Existing Methods}
\label{methods:existing}
Having established dimensions of variance of number representations, we describe some key string-based and real-based methods used in prior work. Table \ref{tab:methods} depicts these methods as individual rows, with the first three columns showing their position in our taxonomy ($\mathsection$ \ref{methods:taxonomy}). The last seven columns correspond to the seven tasks ($\mathsection$ \ref{tasks:existing}), 
with each cell denoting a representative work that introduce it.


\defcitealias{berg-kirkpatrick-spokoyny-2020-empirical}{BS20}
\defcitealias{wallace-etal-2019-nlp}{W+19}
\defcitealias{riedel2018}{SR18}
\defcitealias{geva-etal-2020-injecting}{G+20}
\defcitealias{zhang2020scale}{Z+20}
\defcitealias{jiang-etal-2020-learning}{J+20}
\defcitealias{naik-etal-2019-exploring}{N+19}
\defcitealias{sundararaman-etal-2020-methods}{S+20}
\defcitealias{lin2020nummersense}{L+20}
\defcitealias{goel-etal-2019-pre}{G+19}
\defcitealias{patel2021nlp}{P+21}

\begin{table*}
    \begin{small} 
    \centering
    \begin{tabular}{lccc|ccc|cccc}
         \toprule
         &&&& & \textbf{Exact} & & & \multicolumn{2}{c}{\textbf{Approximate}} & \\
         &&&& \textbf{Arith} & \textbf{Facts} & \textbf{AWP} & \textbf{Num} & \textbf{Mag} & \textbf{Meas} & \textbf{LM$^N$} \\
         \midrule
         \textbf{String-Based} & \textbf{Notation} & \textbf{Tokenization} & \textbf{Pooling} &&&&&&& \\
         Word Vectors & Decimal & Word & NA & \citetalias{wallace-etal-2019-nlp} & & & \citetalias{wallace-etal-2019-nlp} & \citetalias{wallace-etal-2019-nlp} & \citetalias{goel-etal-2019-pre} & \citetalias{jiang-etal-2020-learning} \\
         Contextualized & Decimal & Subword & No & \citetalias{wallace-etal-2019-nlp} & \citetalias{lin2020nummersense} & \citetalias{geva-etal-2020-injecting} & \citetalias{wallace-etal-2019-nlp} & \citetalias{wallace-etal-2019-nlp} & \citetalias{zhang2020scale} & \citetalias{riedel2018} \\
         GenBERT & Decimal & Char & No & \citetalias{geva-etal-2020-injecting} & & \citetalias{geva-etal-2020-injecting} & & & & \\
         NumBERT & Scientific & Char & No & & & & & & \citetalias{zhang2020scale} & \citetalias{zhang2020scale} \\
         DigitRNN/CNN & Decimal & Char & Yes & \citetalias{wallace-etal-2019-nlp} & & & \citetalias{wallace-etal-2019-nlp} & \citetalias{wallace-etal-2019-nlp} & & \citetalias{riedel2018} \\
         DigitRNN-sci & Scientific & Char & RNN & & & & & & & \citetalias{berg-kirkpatrick-spokoyny-2020-empirical} \\
         Exponent & Scientific & Word & NA & & & & & & & \citetalias{berg-kirkpatrick-spokoyny-2020-empirical}  \\
         
         \midrule
         \textbf{Real-Based} & \textbf{Scale} & \textbf{Direction} & \textbf{Binning} &&&&&&& \\
         DICE & Linear & Enc-only & No & \citetalias{sundararaman-etal-2020-methods} & & & \citetalias{sundararaman-etal-2020-methods} & \citetalias{sundararaman-etal-2020-methods} & & \\
         Value & Linear & Both & No & \citetalias{wallace-etal-2019-nlp} & & & \citetalias{wallace-etal-2019-nlp} & \citetalias{wallace-etal-2019-nlp} & & \\
         Log Value & Log & Both & No & \citetalias{wallace-etal-2019-nlp} & & & \citetalias{wallace-etal-2019-nlp} & \citetalias{wallace-etal-2019-nlp} & \citetalias{zhang2020scale} & \\
         MCC & Log & Dec-only & Yes & & & & & & \citetalias{zhang2020scale} & \\
         Log Laplace & Log & Dec-only & No & & & & & & & \citetalias{berg-kirkpatrick-spokoyny-2020-empirical} \\
         Flow Laplace & Learn & Dec-only & No & & & & & & & \citetalias{berg-kirkpatrick-spokoyny-2020-empirical} \\
         DExp & Log & Dec-only & No & & & & & & & \citetalias{berg-kirkpatrick-spokoyny-2020-empirical} \\
         GMM & Linear & Dec-only & Both** & & & & & & & \citetalias{riedel2018}
         \\
         GMM-proto & Linear & Enc-only* & No & \citetalias{jiang-etal-2020-learning} & & & \citetalias{jiang-etal-2020-learning} & \citetalias{jiang-etal-2020-learning} & & \citetalias{jiang-etal-2020-learning} \\
         SOM-proto & Log & Enc-only* & No & \citetalias{jiang-etal-2020-learning} & & & \citetalias{jiang-etal-2020-learning} & \citetalias{jiang-etal-2020-learning} & & \citetalias{jiang-etal-2020-learning} \\
         \bottomrule
    \end{tabular}%
    \caption{
    An overview of numeracy in NLP: Each row is a method ($\mathsection$\ref{methods:existing}), arranged as per our taxonomy ($\mathsection$\ref{methods:taxonomy}) split by string and real, further branching into three dimensions each. The last seven columns correspond to the seven subtasks of numeracy ($\mathsection$\ref{tasks:existing}), split by Exact and Approximate granularity ($\mathsection$\ref{tasks:taxonomy}).
    The cells point to representative (not exhaustive) works that have experimented with a given method (row) on a given task (column).
    Notes:
    Prototype* is encoder-only but reuses embeddings for the decoder \cite{jiang-etal-2020-learning}. GMM** has been discretized \cite{riedel2018} as well as continuous valued \cite{berg-kirkpatrick-spokoyny-2020-empirical}. 
    }
    \label{tab:methods}
    \end{small}
\end{table*}


\subsubsection{String-based methods}

\textbf{Word Vectors \& Contextualized Embeddings}
Word2vec \cite{mikolov2013w2v}, GloVe \cite{penningtonGlove}, ELMo \cite{petersElmo}, and BERT \cite{devlinbert} have been probed as baselines against several contending methods.

\textbf{GenBERT} \citet{geva-etal-2020-injecting} presented GenBERT, a question answering model with pretrained BERT serving as both its encoder and decoder.
GenBERT tokenizes numbers at the digit level, and is finetuned on auxiliary tasks of arithmetic word problems and simple arithmetic.

\textbf{NumBERT}
\citet{zhang2020scale} pretrain BERT from scratch over a modified dataset such that all numbers have been converted into scientific notation, i.e., $314.1$ is expressed as $3141$[EXP]$2$). NumBERT hence follows a scientific notation, subword tokenization, and no pooling.\footnote{\label{pool}Pooling as described in $\mathsection$ \ref{methods:string}.}

\textbf{DigitRNN, DigitCNN} 
\citet{riedel2018} and \citet{wallace-etal-2019-nlp} experimented with pooling
of digit embeddings into a single embedding representing the full number. Both used RNNs as well as CNNs for pooling. 

\textbf{DigitRNN-sci \& Exponent (Embedding)}
\citet{berg-kirkpatrick-spokoyny-2020-empirical} used a scientific notation variant of DigitRNNs (which we refer to as DigitRNN-sci in Table \ref{tab:methods}), as well as a simpler alternative: exponent embedding.
The latter merely learns a lookup embedding for the exponent, completely ignoring the mantissa. 

\subsubsection{Real-based methods}

\textbf{DICE}
Determinisitic Independent-of-Corpus Embeddings \cite{sundararaman-etal-2020-methods} is an attempt to handcraft number encoder \footnote{\label{encdec}Number encoder-decoder as defined in $\mathsection$ \ref{methods:real}.} $f$ so as to preserve the relative magnitude between two numerals and their embeddings. Given two scalars $i$ and $j$, and their embeddings $f(i)$ and $f(j)$, the cosine distance between $f(i)$ and $f(j)$ is intended to monotonically increase/decrease with the Euclidean distance between $i$ and $j$. 
DICE is offered as not only a deterministic encoding but also as an auxiliary loss function for softly training number embeddings alongside, say, SQuAD \cite{rajpurkarSquad}


\textbf{Value Embedding}
The most intuitive parameterized encoder for real numbers is one that feeds the scalar magnitude of the number through a shallow neural network. 
The converse of value embedding is to learn a shallow neural network mapping $g:\mathbb{R}^d\to\mathbb{R}$. This decoder 
is simply the probe used for decoding/numeration task.

The idea of projecting number magnitudes into an NLP model that otherwise inputs only lookup embeddings may appear flawed. But \citet{vaswani2017attention} have (rather successfully) encoded positional information into transformers using both learned embeddings (similar to Value) and fixed ones (similar to DICE). 

\textbf{Log Value}
\citet{wallace-etal-2019-nlp} also experiment with a log-scaled value encoder in addition to the one on a linear scale.
\citet{zhang2020scale} experiment with a log value decoder 
for measurement estimation, which they call the \textbf{RGR} (regress) method.
Log scaling has a neuroscientific inspiration since observations of human (and animal) understanding of numbers is better modelled by a log-scale representation \cite{dehaene2011number}.



\textbf{Log Laplace}
In contrast to the point estimate output of the RGR decoder, models can also be used to parameterize a distribution over numbers. Such a formulation is helpful when estimating approximate quantities. Vectors representing some context can be used to parameterize, say, the mean and variance of a Gaussian or Laplace distribution.  \citet{berg-kirkpatrick-spokoyny-2020-empirical} instead transform the space being modeled by parameterizing the location parameter of a Log-Laplace distribution $L(X,1)$ where $X$ is the context representation of unmasked tokens, in a masked (numerical) language modelling setup. When inferring or decoding a number, they sample a point $z$ \textasciitilde $L(X,1)$ and exponentiate it, such that the output is $exp(z)$. 

\textbf{Flow Laplace} The expressivity of number decoders can be expanded or contracted by merely parameterizing a different distribution. \citet{berg-kirkpatrick-spokoyny-2020-empirical} propose a more expressive decoder where instead of the log scale, the model learns its own density mapping. After sampling $z$ \textasciitilde $L(X,1)$, the output is transformed to $\frac{exp(\frac{z-a}{b})}{c}$, where $a$, $b$, and $c$, are also parameters emitted by the same model.

\textbf{MCC}
or multi-class classification is another number decoder which outputs a distribution, but a discrete one: over log-scaled bins of numbers, e.g., 1-10, 10-100, and so on \cite{zhang2020scale}.
Previously described decoders either output a point estimate or a unimodal distribution, thus failing to hedge its predictions for a multimodal ground truth. Given a masked number prediction problem \emph{We went to the restaurant at [MASK] p.m.}, MCC is better equipped to estimate two peaks: one around lunch time (say, 1-2 p.m.) and another around dinner (say, 7-9 p.m.).

\textbf{Discrete Latent Exponent (DExp)}
is another potentially multimodal distribution \cite{berg-kirkpatrick-spokoyny-2020-empirical} where the model parameterizes a multinomial distribution for the exponent (similar to MCC) and uses it to sample an exponent $e$, which then acts as a latent variable for emitting the mean $\mu$ of a Gaussian (standard deviation fixed at $0.05$). This Gaussian is finally used to sample the output number $z$ \textasciitilde $N(\mu, 0.05)$.

\textbf{GMM}
Another attempt to circumvent the unimodal Gaussians or point estimates is to learn a Gaussian mixture model. \citet{riedel2018} learn a mixture of $K$ Gaussians by pretraining their means ($\mu_i$) and variances (${\sigma_i}^2$) over the training corpus with Expectation Maximization algorithms, while the mixing weights $\pi_i$ are derived from the model. Next, to sample a single number from the GMM probability mass function $q(u) = \sum_{i=1}^{K}{\pi_i N (u; \mu_i; \sigma_i)}$, the authors first sample the precision (number of decimal places) from yet another Gaussian and use that to discretize the probability mass function into equal sized bins, over which the probabilities are summed. If the sampled precision is, say $2$, then the probability of emitting a number $3.14$ is given by $\int_{3.135}^{3.145}{q(u)du}$. This likelihood estimate is used to train a causal language model.

\citet{berg-kirkpatrick-spokoyny-2020-empirical}'s GMM implementation is slightly different: it alters the last inference step by sampling directly from the mixture of Gaussians, as they did with Log Laplace, Flow Laplace, and DExp. 

\textbf{GMM-prototype} by \citet{jiang-etal-2020-learning} similarly pretrains (with EM/hard-EM) the mean, the variances, but also the mixture weights $\pi_i$s of a GMM over the training corpus. They then learn $K$ prototype embeddings $e_i$s corresponding to the $K$ Gaussians. When encoding 
a new numeral $n$, its (input) embedding is calculated as: $E(n) = \sum_{i=1}^K{w_i . e_i}$, where the weights are induced from the GMM:
$$w_i = P(Z=i | U=n) = \frac{\pi_i N(n; \mu_i; \sigma_i)}{\sum_{j=1}^K{\pi_j N(n; \mu_j; \sigma_j)}}$$


Thus the difference between GMM and GMM-prototypes is that after fixing mean and standard deviations of the Gaussian mixtures, in GMM the model learns to predict the mixture weights $\pi_i$ for each individual number prediction, whereas in GMM-prototype, $\pi_i$'s are frozen and the model learns prototype embeddings $e_i$'s. Note that prototype embeddings are encoder-only.
To decode numbers, the authors implement weight-sharing across input and output embeddings, similar to how word vectors are trained \cite{mikolov2013w2v}, i.e., finding out which of the numerals in the corpus has the closest embedding.

\textbf{SOM-prototype}
GMM-prototype, in effect, merely use the mixture of Gaussians to infer prototypes and to get the weights $w_i$'s. \citet{jiang-etal-2020-learning} tried another variant by identifying prototype numerals with Self Organizing Maps \cite{kohonen1990self} and by defining the weights as: 
$w_i = |g(x_i) - g(n)|^{-1}$
where $x_i$ is the $i$th prototype, $n$ is the number to be encoded, and $g$ is a log-based squashing function.


\section{Results}
\label{results}

Having organized the landscape of numeracy tasks and methods, we now present come key results for each numeracy task in NLP from previously published experiments over a subset of the described number representations:

\textbf{Abstract Probes} Word Embeddings vastly outperform random embedding baselines on abstract probes such as numeration, magnitude comparison, and sorting \cite{wallace-etal-2019-nlp,naik-etal-2019-exploring}. 
DICE, Value and Log Value embeddings excel at these probes, which makes intuitive sense given that they explicitly encode the numbers' magnitude - although Value embeddings do not easily extrapolate to larger numbers, possibly due to instability in training. The best number encoders with respect to these probes were found to be DigitCNNs, and character-tokenized models, e.g., ELMo, in general outperform subword ones, e.g., BERT \cite{wallace-etal-2019-nlp}.

\textbf{Arithmetic} GPT-3 \cite{brown2020language} performs extremely well at zero shot simple arithmetic, as long as the number of digits in the operands are low. The tokenization scheme could be the cause for limited extrapolation, since language models get better at arithmetic when numbers are tokenized at the digit/character level \cite{nogueira2021investigating,wallace-etal-2019-nlp}. 
For arithmetic word problems, state of the art solvers rely on predicting an equation, which is then filled in with specific numeric values from the question \cite{patel2021nlp}, altogether bypassing the need for encoding numbers into embeddings. 

\textbf{Masked Language Modelling}
\citet{zhang2020scale} show that BERT pretrained over datasets where numbers are in scientific notation (NumBERT) converges to the same loss as BERT on masked language modelling objective, and scores nearly the same on GLUE language understanding benchmarks. For (causal) numeric language modelling, \citet{riedel2018} show that Gaussian Mixture Models are the best decoders. For (masked) numeric language modelling, \citet{berg-kirkpatrick-spokoyny-2020-empirical} show that modelling the mantissa in scientific notation may be an overkill, since exponent embeddings alone outperform DigitRNN-sci over financial news and scientific articles. 

\textbf{Measurement Estimation}
\citet{zhang2020scale} train a regression probe to predict measurements of objects over the CLS embeddings of BERT/NumBERT. Given a template-lexicalized sentence such as ``the dog is heavy,'' the model must predict the weight of a typical dog, against ground truth from the Distribution over Quantities dataset \cite{elazar-etal-2019-large}. They find that NumBERT is a better text encoder than BERT for measurement estimation, the only difference between them being the notation used by the respective pretraining corpora. They also experiment with two number decoders: MCC (multi-class classification) and RGR (regression / Log Value embedding). MCC performs better when trying to predict Distributions over Quantities - perhaps due to the ground truth resembling the predicted gaussians - but not on VerbPhysics - where the ground truth is less noisy. Lastly, even static word embeddings like GloVe have been shown to contain enough knowledge of measurement estimates to contrast two objects, e.g., classifying whether a \emph{car} is bigger/heavier/fasster than a \emph{ball} \cite{goel-etal-2019-pre}. 

\textbf{Exact Facts} BERT and RoBERTa capture limited numerical commonsense, evident over NumerSense \cite{lin2020nummersense} sentences such as `a tricycle has [MASK] wheels,' with the answer choices limited to the numbers $0$-$10$. Results can be further improved by finetuning over a Wikipedia-extracted dataset of numeric information. \citet{mishra2020question} find the commonsense question answering to be one of the hardest among their Numbergame challenge, using the NumNetv2 model \cite{ran-etal-2019-numnet} which is commonly used for DROP question answering.
Both of these experiments evaluate on exact match metrics, hence it remains to be seen if representing approximate magnitudes yields benefit in modelling numeric facts.

\section{Recommendations}
\label{takeaways}

Based on the above results, we now synthesize certain key insights into a set of directed takeaways to guide practitioners' design of number representations for their task:

\textbf{Rule of thumb for string-based methods?} Scientific notation is superior to decimal notation \cite{zhang2020scale} since models can learn to attend mostly to the exponent embedding rather than the mantissa \cite{berg-kirkpatrick-spokoyny-2020-empirical}. Character level tokenization outperforms subword level \cite{nogueira2021investigating,wallace-etal-2019-nlp,geva-etal-2020-injecting}. Pooled representations (DigitRNN, DigitCNN) lack a controlled study with unpooled ones (NumBERT, GenBERT) which makes it hard to proclaim a winner among the two.

\textbf{Rule of thumb for real-based methods?} 
Log scale is preferred over linear scale \cite{zhang2020scale,jiang-etal-2020-learning,wallace-etal-2019-nlp,berg-kirkpatrick-spokoyny-2020-empirical}, which makes intuitive sense but lacks as rigorous a study as has been undertaken in the cognitive science community \cite{feigenson2004core}. 
Regarding discretization, \citet{zhang2020scale} show that binning (dense cross entropy loss) works better than continuous value prediction (MAE loss) on datasets where ground truth distributions are available.
Lastly, modeling continuous predictions is notoriously hard for large ranges \cite{wallace-etal-2019-nlp} but \citet{riedel2018} offer a way of binning such distributions by picking a precision level.

\textbf{Encoding vs Decoding numbers?} In our simplified discussions above, we avoid differentiating between methods for encoding and decoding numbers. Value Embedding, for instance, can be used to encode numbers (projecting scalars onto vector space) as well as to decode numbers (collapsing a vector into a scalar). On the other hand, manually-designed encoders like DICE are not easily reversible into decoding methods. Even with reversible methods, the encoders and decoders must usually be independently parameterized, unlike the input and output word embeddings which often share weights \cite{press2016using}. Prototype embeddings by \citet{jiang-etal-2020-learning} are an exception, which share input/output embeddings for a fixed vocabulary of numbers.

\textbf{Can we mix-and-match multiple methods?} Given the wide range of number representations, an obvious next step is to try an ensemble of embeddings.
\citet{berg-kirkpatrick-spokoyny-2020-empirical} show that for encoding numbers, exponent embeddings
added to DigitRNN (scientific notation) embeddings barely outperforms the exponent embeddings alone. 
Similar experiments with a mix of real and string methods are yet to be seen.

\textbf{Which methods for which tasks?} Based on our taxonomy of tasks in Table \ref{tab:tasks}, abstract tasks are good early probes for the grounded ones, e.g., finetuning GenBERT \cite{geva-etal-2020-injecting} on simple arithmetic helps it do well on downstream question answering, and the high scores of DICE \cite{sundararaman-etal-2020-methods} on numeration and magnitude comparison are an indicator of similar boosts on (numeric) language modelling.
With respect to granularity, real-based methods work well for approximate tasks such as measurement estimation and language modeling \cite{zhang2020scale,berg-kirkpatrick-spokoyny-2020-empirical} but not for exact tasks like arithmetic word problems or commonsense.
DigitRNNs are broad-purpose number encoders, whereas distribution modeling methods like DExp are effective at decoding numbers.

\section{Vision for Unified Numeracy in NLP}
\label{roadmap}

Numeracy is a core system of human intelligence \cite{kinzler2007core}. Teaching numeracy to students works best when taught holistically, while less effective teachers deal with areas of mathematics discretely \cite{askew1997effective}. While the NLP community has genuinely strived to improve language models' numeric skills, not all aspects of numeracy have been sufficiently targeted. 
It is evident from the sparsity in Table \ref{tab:methods} that the community is far from achieving, or attempting, a holistic solution to numeracy. In this section, we outline our vision for such a unified solution, as three prerequisites necessary to consider for numerical NLU:

\textbf{Evaluation.} The first step towards a holistic solution to numeracy requires a benchmark covering its different subtasks. Aggregated leaderboards in NLP like GLUE \cite{wang2018glue} and SuperGLUE \cite{wang2019superglue} have incentivized research on natural langauge understanding, with scores categorized into semantic, syntactic, logical, and background knowledge.

An analogous leaderboard could be constructed to evaluate models on numeric reasoning tasks, again categorized according to the skills evaluated, e.g., exact vs approximate granularity, or abstract vs grounded numeracy. Numbergame \cite{mishra2020question} is one such aggregation focusing on exact numeracy benchmarks, as evaluated by F1 and exact match scores in a reading comprehension setup.
Both Numbergame and our own list of tasks (Section \ref{tasks:existing}) are preliminary attempts at teasing apart the different aspects of numeracy. We encourage researchers to extend and refine such taxonomies.

A suite of numeracy tasks, matched with evaluations of their respective numerical skills, can enable testing model generalization from one skill to another.
Some progress has already been made in this transfer learning setup, e.g., GenBERT \cite{geva-etal-2020-injecting}, finetuned on a synthetic dataset of arithmetic problems, is found to score higher on DROP QA. Similarly, DICE \cite{sundararaman-etal-2020-methods}, optimized for numeration, improves score on Numeracy600K order-of-magnitude prediction task. Going forward, we need several such studies, ideally for each pair of tasks to see whether some numeracy skills help models generalize to others.

\textbf{Design Principles.} 
Number representations vary based on design trade-offs between inductive biases and data-driven variance.
The default BERT setup, with subword tokenization and lookup embeddings, occupies the variance end of the spectrum, allowing freedom in representing numbers. Value embeddings and DICE encodings, on the other hand, are closer to the bias end of the spectrum, since the inductive bias of continuity on the number line constrains the learning space. It is important to identify where on the bias-variance scale any representation stands, for a fair comparison. 

Following parallel work in cognitive science, the community could explore whether exact and approximate numeracy require two specialized modules \cite{feigenson2004core} or could be handled with a single representation \cite{cordes2001variability}.

Model designers must also make a choice on coverage: whether to target a broad or a narrow range of numbers to be represented. Multi-class classification \cite{zhang2020scale} over a fixed number of bins, restricts the range of numbers expressed, as do DICE embeddings \cite{sundararaman-etal-2020-methods}. Value embeddings are continuous and theoretically unrestricted, but must practically be capped for bug-free training. On the other hand, string-based representations could always fall back to subword/char-level token embeddings to represent not only floats but also irrational ($\sqrt{2}$) and complex ($1+2\iota$) numbers. \citet{royNLI} introduced the Quantity-Value Representation format to allow closed and open ranges alongside scalar point numbers.

\textbf{Broader Impact.} Numbers are ubiquitous in natural language and are easily identified, at least in numeral forms. But they are by no means the only class of ordered concepts required for natural language understanding. Successful number representations can inspire work on incorporating more continuous domains into natural language processing systems. For instance, gradable adjectives like good, great, amazing, etc. are arguably on some cardinal scale, which can be mapped using value embeddings or Gaussian mixture models \cite{sharp-etal-2018-grounding,de-marneffe-etal-2010-good}. Days of the week (Mon-Sun) and months of an year (Jan-Dec) form periodic patterns which can be modeled with sinusoidal functions \cite{martinezWeCNLP}.

Lastly, numeracy is essential for natural language understanding.
Consider the sentence: ``\emph{Programmers earn \$200,000 versus \$100,000 for researchers.}" An intelligent agent with numeracy skills would identify that \$100k is half of \$200k, that \$100k possibly denotes annual salary, and infer that higher salaries lead to higher standards of living. In short, it was able to learn something about the two concepts \emph{programmers} and \emph{researchers}, by crossing the continuous semantic space of numbers! The agent could now make use of this knowledge in a number-free situation, e.g., the mask in ``\emph{He could not afford a car for several years after earning a CS degree because she took a job as a [MASK]}'' might better be filled with the word \emph{researcher}, than with \emph{programmer}.
A key goal of imparting numeracy to NLP models is to help them understand more about the world, \emph{using} numbers. 




\section{Conclusion}
\label{conclusion}

This paper summarizes and contextualizes recent work on numeracy in NLP. We propose the first taxonomy of tasks and methods concerning text-centric numeric reasoning. We highlight key takeaways from the several experiments in literature, along with caveats and scope for confirming some of the observed trends. We present a case for lack of a holistic solution to numeracy in NLP, and put forward a set of aspects to consider when working towards one. 
We draw the following two major conclusions from our study: (1) the default subword segmentation with lookup embeddings used to represent words is clearly suboptimal for numbers (2) there are several unanswered research questions on the level of specificity, coverage, and inductive bias needed to holistically solve numeracy.
\section{Acknowledgements}
\label{ack}

This work was funded by the Defense Advanced Research Projects Agency with award N660011924033. We would like to thank the countless suggestions we accumulated during preliminary presentations at MLSS 2020, WeCNLP 2020, and GSS 2020, as well as over email correspondences with Biplav Srivastava, Antoine Bosselut, and Harsh Agarwal. We would like to thank the anonymous NAACL 2021 reviewers (particularly \#3) for pointing out blind spots in our submission, which we have tried our best to rectify.

\section*{Ethical Considerations}
\label{ethics}
This work revolves around the Hindu-Arabic Numeral system and English number words, which are not the only number systems still in use today. We encourage follow-up work to take these systems into consideration, on the lines of \citet{johnson-etal-2020-probing} and \citet{nefedov2020dataset}.



\bibliography{naacl2021.bib}
\bibliographystyle{acl_natbib}

\appendix


\appendix
\section{Other Numeracy Tasks}
\label{tasks:misc}
Here, we describe certain related tasks that fall outside our taxonomy:

\label{sec:appendix}
\textbf{(Numeric) Paraphrasing} is what we call the task of identifying one-to-one correspondences between different surface forms of the same number. Twelve is the same as `12', also referred to as a dozen. 
This task cuts across all the tasks we discussed, since the same number, expressed in several different ways, should be nevertheless identified by an NLP model before any subsequent reasoning. Similar to how WordNet \cite{miller1995wordnet} provides a huge list of synonyms, numeric paraphrases can be obtained by libraries\footnote{Example: https://pypi.org/project/num2words/} which convert numerals to words, words to numerals, etc. One could also envision this as a learning task given a large enough corpus, such as the NumGen dataset \cite{williams-power-2010-fact} containing 2000 fact-aligned numeric expressions over 110 articles.

\textbf{Quantity Entailment} tasks \cite{ravichanderEquate,royNLI} are analogous to Natural Language Inference, which requires understanding of not only equivalence (as in paraphrasing) but also deeper relations like entailment and contradiction, e.g., the premise `he was 16 yrs old' entails the hypothesis `he was a teenager'. On similar lines, \citet{mishra2020question} modified the existing QuaRel dataset \cite{tafjord2019quarel} to force models to perform quantity entailment, e.g., \emph{dog1 is light, dog2 is heavy} is replaced with \emph{dog1 weighs 70 lbs, dog2 weighs 90 lbs}.

\textbf{Numeral Understanding} is the task of categorizing numbers into percentages, prices, dates, times, quantities, etc. and their respective subcategories \cite{chen2018numeral}.

\textbf{Fused-Head Resolution} for numbers is essential to ground them when the context is implicit. For example, the sentence ``\emph{I woke up at 11}" has `a.m.' or `o'clock' as the fused head to be resolved \cite{elazarHead}.

\textbf{Counting} is the task of keeping track of discrete instances of some object. When kids count a set of objects, they quickly learn to keep a track, say on their fingers, but struggle with realizing the Cardinal Principle, i.e., the last counter value denotes the number of entities being considered \cite{wynn1990children}. Similarly, LSTMs \cite{suzgunLstm} and transformers \cite{bhattamishraTransformers} have been shown to possess counting skills but in order to answer counting questions, they must also learn to map the counts to number words or numerals.
Counting tasks have been proposed in computer vision \cite{testolin2020visual} as well as in NLP \cite{postmaSemeval,talmor2020leap}.



\textbf{Domain-specific} tasks require background knowledge in addition to mathematical skills. Numbergame \cite{mishra2020question} includes questions on Physics (\emph{find the distance travelled in 2 hrs by a train moving at 50 mph}) and Chemistry (\emph{find the mass percentage of H in C6H6}). Project Aristo \cite{clark2019f} solves elementary and high school science problems, which often involve numeric reasoning.


\end{document}